\documentclass{sig-alternate}
\usepackage{algorithm,algorithmic}
\usepackage{amssymb, multirow, paralist}
\usepackage{graphicx,color}

\newtheorem{thm}{Theorem}
\newtheorem{prop}{Proposition}

\def \x {\mathbf{x}}
\def \R {\mathbb{R}}

\def \L {\mathcal{L}}

\begin{document}
%
% --- Author Metadata here ---
%\conferenceinfo{WOODSTOCK}{'97 El Paso, Texas USA}
%\CopyrightYear{2007} % Allows default copyright year (20XX) to be over-ridden - IF NEED BE.
%\crdata{0-12345-67-8/90/01}  % Allows default copyright data (0-89791-88-6/97/05) to be over-ridden - IF NEED BE.
% --- End of Author Metadata ---

\title{Efficient Distance Metric Learning by Adaptive Sampling and Mini-Batch Stochastic Gradient Descent (SGD)}
%\subtitle{[Extended Abstract]
%\titlenote{A full version of this paper is available as
%\textit{Author's Guide to Preparing ACM SIG Proceedings Using
%\LaTeX$2_\epsilon$\ and BibTeX} at
%\texttt{www.acm.org/eaddress.htm}}}
%
% You need the command \numberofauthors to handle the 'placement
% and alignment' of the authors beneath the title.
%
% For aesthetic reasons, we recommend 'three authors at a time'
% i.e. three 'name/affiliation blocks' be placed beneath the title.
%
% NOTE: You are NOT restricted in how many 'rows' of
% "name/affiliations" may appear. We just ask that you restrict
% the number of 'columns' to three.
%
% Because of the available 'opening page real-estate'
% we ask you to refrain from putting more than six authors
% (two rows with three columns) beneath the article title.
% More than six makes the first-page appear very cluttered indeed.
%
% Use the \alignauthor commands to handle the names
% and affiliations for an 'aesthetic maximum' of six authors.
% Add names, affiliations, addresses for
% the seventh etc. author(s) as the argument for the
% \additionalauthors command.
% These 'additional authors' will be output/set for you
% without further effort on your part as the last section in
% the body of your article BEFORE References or any Appendices.

\numberofauthors{5} %  in this sample file, there are a *total*
% of EIGHT authors. SIX appear on the 'first-page' (for formatting
% reasons) and the remaining two appear in the \additionalauthors section.
%
\author{
% You can go ahead and credit any number of authors here,
% e.g. one 'row of three' or two rows (consisting of one row of three
% and a second row of one, two or three).
%
% The command \alignauthor (no curly braces needed) should
% precede each author name, affiliation/snail-mail address and
% e-mail address. Additionally, tag each line of
% affiliation/address with \affaddr, and tag the
% e-mail address with \email.
%
% 1st. author
%\alignauthor
Qi Qian$^\dagger$, Rong Jin$^\dagger$, Jinfeng Yi$^\dagger$, Lijun Zhang$^\dagger$ and Shenghuo Zhu$^\ddagger$\\
\affaddr{$^\dagger$Department of Computer Science and Engineering}\\
\affaddr{Michigan State University, East Lansing, MI, 48824, USA}\\
\affaddr{$^\ddagger$NEC Laboratories America, Cupertino, CA, 95014, USA}\\
\email{\{qianqi, rongjin, yijinfen, zhanglij\}@cse.msu.edu, zsh@nec-labs.com}\\
       }
% There's nothing stopping you putting the seventh, eighth, etc.
% author on the opening page (as the 'third row') but we ask,
% for aesthetic reasons that you place these 'additional authors'
% in the \additional authors block, viz.
%\additionalauthors{Additional authors: John Smith (The Th{\o}rv{\"a}ld Group,
%email: {\texttt{jsmith@affiliation.org}}) and Julius P.~Kumquat
%(The Kumquat Consortium, email: {\texttt{jpkumquat@consortium.net}}).}
%\date{30 July 1999}
% Just remember to make sure that the TOTAL number of authors
% is the number that will appear on the first page PLUS the
% number that will appear in the \additionalauthors section.

\maketitle
\begin{abstract}

Distance metric learning (DML) is an important task that has found applications in many domains. The high computational cost of DML arises from the large number of variables to be determined and the constraint that a distance metric has to be a positive semi-definite (PSD) matrix. Although stochastic gradient descent (SGD) has been successfully applied to improve the efficiency of DML, it can still be computationally expensive because in order to ensure that the solution is a PSD matrix, it has to, at {\it every iteration}, project the updated distance metric onto the PSD cone, an expensive operation. We address this challenge by developing two strategies within SGD, i.e. mini-batch and adaptive sampling, to effectively reduce the number of updates (i.e., projections onto the PSD cone) in SGD. We also develop hybrid approaches that combine the strength of adaptive sampling with that of mini-batch online learning techniques to further improve the computational efficiency of SGD for DML. We prove the theoretical guarantees for both adaptive sampling and mini-batch based approaches for DML. We also conduct an extensive empirical study to verify the effectiveness of the proposed algorithms for DML.

\end{abstract}

% A category with the (minimum) three required fields
%A category including the fourth, optional field follows...
%\category{D.2.8}{Software Engineering}{Metrics}[complexity measures, performance measures]


 % NOT required for Proceedings

\section{Introduction}

%%%%%%%%%%%%%%%%%%%%%%%%%%%%%%%%%%%%%%%%%%%%%%%%%%%%%%%%%%%%%%%%%%
%introduce distance metric learning
%%%%%%%%%%%%%%%%%%%%%%%%%%%%%%%%%%%%%%%%%%%%%%%%%%%%%%%%%%%%%%%%%%

Distance metric learning (DML) is an important subject, and has found applications in many domains, including information retrieval~\cite{HeMZ04}, supervised classification~\cite{weinberger2009}, clustering~\cite{XingNJR02}, and semi-supervised clustering~\cite{ChangY04}. The objective of DML is to learn a distance metric consistent with a given set of constraints, namely minimizing the distances between pairs of data points from the same class and maximizing the distances between pairs of data points from different classes. The constraints are often specified in the form of must-links, where data points belong to the same class, and cannot-links, where data points belong to different classes. The constraints can also be specified in the form of triplets $(\x_i, \x_j, \x_k)$~\cite{weinberger2009}, in which $\x_i$ and $\x_j$ belong to a class different from that of $\x_k$ and therefore $\x_i$ and $\x_j$ should be separated by a distance smaller than that between $\x_i$ and $\x_k$. In this work, we focus on DML using triplet constraints due to its encouraging performance~\cite{chechik2010,shaw2011,weinberger2009}.

%%%%%%%%%%%%%%%%%%%%%%%%%%%%%%%%%%%%%%%%%%%%%%%%%%%%%%%%%%%%%%%%%%
%Challenge of DML
%%%%%%%%%%%%%%%%%%%%%%%%%%%%%%%%%%%%%%%%%%%%%%%%%%%%%%%%%%%%%%%%%%

The main computational challenge in DML arises from the restriction that the learned distance metric must be a positive semi-definite (PSD) matrix, which is often referred as the {\it PSD constraint}. Early approach~\cite{XingNJR02} addressed the PSD constraint by exploring the technique of semi-definite programming (SDP)~\cite{boyd2004convex}, which unfortunately does not scale to large and high dimensional datasets. More recent approaches~\cite{chechik2010,shaw2011} addressed this challenge by exploiting the techniques of online learning and stochastic optimization, particularly stochastic gradient descent (SGD), that only needs to deal with one constraint at each iteration. Although these approaches are significantly more efficient than the early approach, they share one common drawback: in order to ensure that the learned distance metric is PSD, these approaches require, {\it at each iteration}, projecting the updated distance metric onto the PSD cone. The projection step requires performing the eigen-decomposition for a given matrix, and therefore is computationally expensive~\footnote{The computational cost is $O(d^2)$ if we only need to compute the top eigenvectors of the distance metric and becomes $O(d^3)$ if all the eigenvalues and eigenvectors have to be computed for the projection step, where $d$ is the dimensionality of the data.}. As a result, the key challenge in developing efficient SGD algorithms for DML is how to reduce the number of projections without affecting the performance of DML.

%%%%%%%%%%%%%%%%%%%%%%%%%%%%%%%%%%%%%%%%%%%%%%%%%%%%%%%%%%%%%%%%%%
%Mini-batch and our methods
%%%%%%%%%%%%%%%%%%%%%%%%%%%%%%%%%%%%%%%%%%%%%%%%%%%%%%%%%%%%%%%%%%

%\begin{figure*}[t]
%\centering
%\includegraphics[width = 6in]{1.jpg}
%\caption{The main procedure of {\bf SGD combined with mini-batch and adaptive sampling}}\label{fig1}
%\end{figure*}

A common approach for reducing the number of updates and projections in DML is to use the non-smooth loss function. A popular choice of the non-smooth loss function is the hinge loss, whose derivative becomes zero when the input value exceeds a certain threshold. Many online learning algorithms for DML~\cite{chechik2010,DavisKJSD07,JainKDG08} take advantage of the non-smooth loss function to reduce the number of updates and projections. In~\cite{shaw2011}, the authors proposed a structure preserving metric learning algorithm (SPML) that combines a mini-batch strategy with the hinge loss to further reduce the number of updates for DML. It groups multiple constraints into a mini-batch and performs only one update of the distance metric for each mini-batch. But, according to our empirical study, although SPML reduces the running time of the standard SGD algorithm, it results in a significantly worse performance for several datasets, due to the deployment of the mini-batch strategy.

In this work, we first develop a new mini-batch based SGD algorithm for DML, termed {\bf Mini-SGD}. Unlike SPML that relies on the hinge loss, the proposed Mini-SGD algorithm uses a {\it smooth} loss function for DML. We show theoretically that by using a smooth loss function, Mini-SGD is able to achieve similar convergence rate as the standard SGD algorithm but with significantly less number of updates. The second contribution of this work is to develop a new strategy, termed {\bf adaptive sampling}, for reducing the number of projections in DML. The key idea of adaptive sampling is to first measure the ``difficulty'' in classifying a constraint using the learned distance metric, and then perform stochastic updating based on the classification difficulty. More specifically, given the distance metric $M_t$ and triplet $(\x_i^t, \x_j^t, \x^t_k)$, we first measure the difficulty in classifying the triplet $(\x_i^t, \x_j^t, \x^t_k)$ by $\gamma_t = \ell'(\x_i^t, \x_j^t, \x^t_k; M_t)$, where $\ell(\x_i^t, \x_j^t, \x^t_k; M_t)$ is the loss function that measures the classification error. We then sample a binary variable $Z_t$ with $\Pr(Z_t = 1) \propto \gamma_t$, and only update the distance metric when $Z_t = 1$. We refer to the proposed approach for DML as {\bf AS-SGD} for short. Finally, we develop two {\bf hybrid approaches}, termed {\bf HA-SGD} and {\bf HR-SGD}, that combine adaptive sampling with mini-batch to further improve the computational efficiency of SGD for DML. We conduct an extensive empirical study to verify the effectiveness and efficiency of the proposed algorithms for DML.

%%%%%%%%%%%%%%%%%%%%%%%%%%%%%%%%%%%%%%%%%%%%%%%%%%%%%%%%%%%%%%%%%%
%Contribution
%%%%%%%%%%%%%%%%%%%%%%%%%%%%%%%%%%%%%%%%%%%%%%%%%%%%%%%%%%%%%%%%%%

%In summary, the main contributions of this work are:
%\begin{itemize}
%\item Despite the encouraging results for empirical studies~\cite{shaw2011}, the authors did not present a complete theoretical justification for mini-batch DML. We fill out this gap by developing performance guarantee for mini-batch DML. Our analysis significantly improves the previous theoretical result for mini-batch based SGD~\cite{cotter2011}.
%\item We develop an adaptive sampling strategy for efficient DML that dramatically reduces the number of projection steps in SGD. We verify the effectiveness of adaptive sampling based SGD for DML both theoretically and empirically.
%\item We combine adaptive sampling with mini-batch to further reduce the number of projection steps in SGD for DML. We show that the combined approach can be more effective than either adaptive sampling or mini-batch based SGD for DML.
%\end{itemize}

The rest of the paper is organized as follows: Section~\ref{sec:related-work} reviews the related work on distance metric learning and stochastic gradient descent with reduced number of projection steps. Section~\ref{sec:alg} describes the proposed SGD algorithms for DML based on mini-batch and adaptive sampling. Two hybrid approaches are presented that combine mini-batch and adaptive sampling for DML. The theoretical guarantees for both mini-batch based and adaptive sampling based SGD are also presented in Section~\ref{sec:alg}. Section~\ref{sec:exp} summarizes the results of the empirical study, and Section~\ref{sec:conclusion} concludes this work with future directions.

\section{Related work}
\label{sec:related-work}

%%%%%%%%%%%%%%%%%%%%%%%%%%%%%%%%%%%%%%%%%%%%%%%%%%%%%%%%%%%%%%%%%%
%General DML
%%%%%%%%%%%%%%%%%%%%%%%%%%%%%%%%%%%%%%%%%%%%%%%%%%%%%%%%%%%%%%%%%%
Many algorithms have been developed to learn a linear distance metric from pairwise constraints, where must-links include pairs of data points from the same class and cannot-links include pairs of data points from different classes (~\cite{liu2006} and references therein). Besides pairwise constraints, an alternative strategy is to learn a distance metric from a set of triplet constraints $(\x_i^t, \x_j^t, \x_k^t), t=1, \ldots, N$, where $\x_i^t$ is expected to be closer to $\x_j^t$ than to $\x_k^t$. Previous studies~\cite{chechik2010,shaw2011,weinberger2009} showed that triplet constraints could be more effective for DML than pairwise constraints.

%%%%%%%%%%%%%%%%%%%%%%%%%%%%%%%%%%%%%%%%%%%%%%%%%%%%%%%%%%%%%%%%%%
%Online DML
%%%%%%%%%%%%%%%%%%%%%%%%%%%%%%%%%%%%%%%%%%%%%%%%%%%%%%%%%%%%%%%%%%

Several online algorithms have been developed to reduce the computational cost of DML~\cite{chechik2010,DavisKJSD07,GlobersonR05,JainKDG08}. Most of these methods are based on stochastic gradient descent. At each iteration, they randomly sample {\it one} constraint, and update the distance metric based on the sampled constraint. The updated distance metric is further projected onto the PSD cone to ensure that it is PSD. Although these approaches are significantly more scalable than the batch learning algorithms for DML~\cite{weinberger2009}, they suffer from the high computational cost in the projection step that has to be performed at {\it every} iteration. A common approach for reducing the number of projections is to use a non-smooth loss function, such as the hinge loss. In addition, in~\cite{shaw2011}, the authors proposed a structure preserving metric learning (SPML) that combines mini-batch with the hinge loss to further reduce the number of projections. The main problem with the approach proposed in~\cite{shaw2011} is that according to the theory of mini-batch, it only works well with a smooth loss. Since the hinge loss is a non-smooth loss function, combining mini-batch with the hinge loss may result in a suboptimal performance. This is verified by our empirical study in which we observed that the distance metric learned by SPML performs significantly worse than that learned by the standard stochastic gradient descent method. We resolve this problem by presenting a new SGD algorithm for DML that combines mini-batch with a smooth loss, instead of the hinge loss.

%%%%%%%%%%%%%%%%%%%%%%%%%%%%%%%%%%%%%%%%%%%%%%%%%%%%%%%%%%%%%%%%%%
%Projection free and projection one
%%%%%%%%%%%%%%%%%%%%%%%%%%%%%%%%%%%%%%%%%%%%%%%%%%%%%%%%%%%%%%%%%%

Finally, it is worthwhile mentioning several recent studies proposed to avoid projections in SGD. In~\cite{hazan2012}, the authors developed a projection free SGD algorithm that replaces the projection step with a constrained linear programming problem. In~\cite{mahdavi2012}, the authors proposed a SGD algorithm with only one projection that is performed at the end of the iterations. Unfortunately, the improvement of the two algorithms in computational efficiency is limited, because they require computing,  {\it at each iteration}, the minimum eigenvalue and eigenvector of the updated distance metric, an operation with $O(d^2)$ cost, where $d$ is the dimensionality of the data.

\section{Improved SGD for DML by Mini-batch and Adaptive Sampling}
\label{sec:alg}
%%%%%%%%%%%%%%%%%%%%%%%%%%%%%%%%%%%%%%%%%%%%%%%%%%%%%%%%%%%%%%%%%%
%introduction
%%%%%%%%%%%%%%%%%%%%%%%%%%%%%%%%%%%%%%%%%%%%%%%%%%%%%%%%%%%%%%%%%%
We first review the basic framework of DML with triplet constraints. We then present two strategies to improve the computational efficiency of SGD for DML, one by mini-batch and one by adaptive sampling. We present the theoretical guarantees for both strategies, and defer more detailed analysis to the appendix. At the end of this section, we present two hybrid approaches that combine mini-batch with adaptive sampling for more efficient DML.

\subsection{DML with Triplet Constraints}
%%%%%%%%%%%%%%%%%%%%%%%%%%%%%%%%%%%%%%%%%%%%%%%%%%%%%%%%%%%%%%%%%%
%formulation
%%%%%%%%%%%%%%%%%%%%%%%%%%%%%%%%%%%%%%%%%%%%%%%%%%%%%%%%%%%%%%%%%%
Let $\mathcal{X} \subset \R^d$ be the domain for input patterns, where $d$ is the dimensionality. For the convenience of analysis, we assume all the input patterns with bounded norm, i.e. $\forall \x \in \mathcal{X}, |\x|_2 \leq r$. Given a distance metric $M \in \R^{d\times d}$, the distance square between $\x_a$ and $\x_b$, denoted by $|\x_a - \x_b|^2_M$, is measured by
\[
|\x_a - \x_b|_M^2 = (\x_a - \x_b)^{\top}M(\x_a - \x_b)
\]
Let $\Omega = \left\{M: M\succeq 0, \|M\|_F \leq R\right\}$ be the domain for distance metric $M$, where $R$ specifies the domain size. Let $\mathcal{D} = \{(\x_i^1, \x_j^1, \x^1_k), \ldots, (\x_i^N, \x_j^N, \x^N_k)\}$ be the set of triplet constraints used for DML, where $\x^t_i$ is expected to be closer to $\x^t_j$ than to $\x^t_k$. Let $\ell(z)$ be the convex loss function. Define $\Delta(\x_i^t, \x_j^t, \x_k^t; M)$ as
\begin{eqnarray*}
\lefteqn{\Delta(\x_i^t, \x_j^t, \x_k^t; M) = |\x^t_i - \x^t_k|_M^2 - |\x^t_i - \x^t_j|_M^2} \\
& = & \left\langle M, (\x^t_i - \x^t_k)(\x^t_i - \x^t_k)^{\top} - (\x^t_i - \x^t_j)(\x^t_i - \x^t_j)^{\top}\right\rangle \\
& = & \langle M, A_t\rangle
\end{eqnarray*}
where
\[
    A_t = (\x^t_i - \x^t_k)(\x^t_i - \x^t_k)^{\top} - (\x^t_i - \x^t_j)(\x^t_i - \x^t_j)^{\top}
\]
Given the triplet constraints in $\mathcal{D}$ and the domain in $\Omega$, we learn an optimal distance metric $M \in \R^{d\times d}$ by solving the following optimization problem
\begin{eqnarray}
& \min\limits_{M \in \Omega} & \L(M) = \frac{1}{N}\sum_{t=1}^N \ell\left(\Delta(\x_i^t, \x_j^t, \x_k^t; M) \right) \label{eqn:opt}
\end{eqnarray}
The key idea of online DML is to update the distance metric based on one sampled constraint at each iteration. More specifically, at iteration $t$, it samples a triplet constraint $(\x_i^t, \x_j^t, \x_k^t)$, and updates the distance metric $M_t$ to $M_{t+1}$ by
\[
M_{t+1} = \Pi_{\Omega}\left(M_t - \eta\ell'(\Delta(\x_i^t, \x_j^t, \x_k^t; M_t))A_t\right)
\]
where $\eta > 0$ is the step size, $\ell'(\cdot)$ is the derivative and $\Pi_{\Omega}(M)$ projects a matrix $M$ onto the domain $\Omega$. The following proposition shows $\Pi_{\Omega}(M)$ can be computed in two steps, i.e. first projecting $M$ onto the PSD cone, and then scaling the projected $M$ to fit in with the constraint $\|M\|_F \leq R$.
\begin{prop}~\cite{boyd2004convex} \label{prop:1}
We have
\[
    \Pi_{\Omega}(M) = \frac{1}{\max(\|M'\|_F/R, 1)} M'
\]
where $M' = P(M)$ and $P(M)$ projects matrix $M$ onto the PSD cone.
\end{prop}
As indicated by Proposition~\ref{prop:1}, $\Pi_{\Omega}(M)$ requires projecting distance metric $M$ onto the PSD cone, an expensive operation that requires eigen-decomposition of $M$.

Finally, to bound both the regret and the number of updates, in this study, we approximate the hinge loss by a smooth loss function
\begin{eqnarray}
    \ell(z) = \frac{1}{L}\log(1 + \exp\left(-L(z - 1) \right)) \label{eqn:smooth}
\end{eqnarray}
where $L > 0$ is a parameter that controls the approximation error: the larger the $L$, the closer $\ell(z)$ is to the hinge loss. Note that the smooth approximation of the hinge loss was first suggested in~\cite{zhang2001} for classification and was later verified by an empirical study in~\cite{ZhangJYH03}. The key properties of the loss function $\ell(z)$ in (\ref{eqn:smooth}) are given in the following proposition.
\begin{prop} For the loss function defined in (\ref{eqn:smooth}), we have
\[
\forall z \in \R, \quad |\ell'(z)| \leq 1, \; |\ell'(z)| \leq L\ell(z)
\]
\end{prop}
Compared to the hinge loss function, the main advantage of the loss function in (\ref{eqn:smooth}) is that it is a smooth loss function. As will be revealed by our analysis, it is the smoothness of the loss function that allows us to effectively explore both the mini-batch and adaptive sampling strategies for more efficient DML without having to sacrifice the prediction performance.

\subsection{Mini-batch SGD for DML (Mini-SGD)}
%%%%%%%%%%%%%%%%%%%%%%%%%%%%%%%%%%%%%%%%%%%%%%%%%%%%%%%%%%%%%%%%%%
%mini-batch
%%%%%%%%%%%%%%%%%%%%%%%%%%%%%%%%%%%%%%%%%%%%%%%%%%%%%%%%%%%%%%%%%%
Mini-batch SGD improves the computational efficiency of online DML by grouping multiple constraints into a mini-batch and only updating the distance metric once for each mini-batch. For brevity, we will refer to this algorithm as {\bf Mini-SGD} in the rest of the paper.

Let $b$ be the batch size. At iteration $t$, it samples $b$ triplet constraints, denoted by
\[
(\x_i^{t,s}, \x_j^{t,s}, \x_k^{t,s}), s= 1,\ldots, b,
\]
and defines the mini-batch loss at iteration $t$ as
\[
\ell_t(M_t) = \frac{1}{b}\sum_{s=1}^b \ell\left(\Delta(\x_i^{t,s}, \x_j^{t,s}, \x_k^{t,s}; M_t)\right)
\]
Mini-batch DML updates the distance metric $M_t$ to $M_{t+1}$ using the gradient of the mini-bach loss function $\ell_t(M)$, i.e.,
\[
M_{t+1} = \Pi_{\Omega}\left(M_t - \eta\nabla \ell_t(M_t)\right)
\]
Algorithm~\ref{alg:1} gives the detailed steps of Mini-SGD for DML, where step 5 uses Proposition 1 for computing the projection $\Pi_{\Omega}(\cdot)$.

\begin{algorithm}[t]
\caption{Mini-batch Stochastic Gradient Descent (Mini-SGD) for DML}
\begin{algorithmic}[1]

\STATE {\bf Input:} triplet constraints $\{(\x_i^t, \x_j^t, \x_k^t)\}_{t=1}^N$, step size $\eta$, mini-batch size $b$, and domain size $R$

\STATE Initialize $M_1 = I$ and $T = N/b$

\FOR{$t=1, \ldots, T$}
    \STATE Sample $b$ triplet constraints $\{(\x_i^{t,s}, \x_j^{t,s}, \x_k^{t,s})\}_{s=1}^b$
    \STATE Update the distance metric by
        \begin{eqnarray*}
            M_{t+1} &=& \Pi_{\Omega}\left(M_t - \eta\nabla\ell_t(M_t)\right)
 %           M_{t+1}&=& \frac{1}{\max\left(\|\Mh_{t+1}\|_F/R,1\right)} \Mh_{t+1}
        \end{eqnarray*}
\ENDFOR
\RETURN $\bar{M} = \frac{1}{T}\sum_{t=1}^{T} M_{t}$
\end{algorithmic}\label{alg:1}
\end{algorithm}

The theorem below provides the theoretical guarantee for the Mini-SGD algorithm for DML using the smooth loss function defined in (\ref{eqn:smooth}).
\begin{thm}\label{thm:mini-batch}
Let $\bar{M}$ be the solution output by Algorithm~\ref{alg:1} that uses the loss function defined in (\ref{eqn:smooth}). Let $M_*$ be the optimal solution to (\ref{eqn:opt}). Assume $\|A_t\|_F \leq A$ for any triplet constraint. For a fixed $\delta \in (0, 1)$, we have, with a probability $1 - 2\delta$:
\begin{eqnarray}
\lefteqn{\L(\bar{M}) \leq \frac{\L(M_*)}{1 -  3\eta LA^2} + \frac{b R^2}{2(1 - 3\eta LA^2)\eta N}} \nonumber \\
& & + \frac{C_1 A^2 \eta}{(1 - 3\eta LA^2)N}\left[\log\frac{2N}{\delta b}\right]^2\log\frac{m}{\delta} \label{eqn:mini-bound}
\end{eqnarray}
where $m = \lceil \log_2 N \rceil$, and $C_1$ is an universal constant that is at most $32$.
\end{thm}
Figure~\ref{fig1} shows the reduction in the training error over the number of triplet constraints by the Mini-SGD algorithm on three datasets~\footnote{The information of these datasets can be found in the experimental section.}. Compared to the standard SGD algorithm, we observe that Mini-SGD converges to a similar value of training error, thus validating our theorem empirically.

\noindent{\bf Remark 1} We observe that the second term in the upper bound in (\ref{eqn:mini-bound}), i.e., $bR^2/[2(1 - 3\eta LA^2)\eta N]$, has a linear dependence on mini-batch size $b$, implying that the larger the $b$, the less accurate the distance metric learned by Algorithm~\ref{alg:1}. Hence, by adjusting parameter $b$, the size of mini-batch, we are able to make appropriate tradeoff between the prediction accuracy and the computational efficiency: the smaller the $b$, the more accurate the distance metric but with more updates and consequentially higher computational cost. Finally, it is worthwhile comparing Theorem~\ref{thm:mini-batch} to the theoretical result for a general mini-batch SGD algorithm given in~\cite{cotter2011}, i.e.
\begin{eqnarray}\label{eqn:minibound}
    \L(\bar{M}) \leq \L(M_*) + O\left(\frac{1}{\sqrt{N}} + \frac{b^2}{N^2} \right) \label{eqn:bound-2}
\end{eqnarray}
It is clear that Theorem~\ref{thm:mini-batch} gives a significantly better result when the optimal loss $\L(M_*)$ is small (i.e. when the triplet constraints can be well classified by the optimal distance metric $M_*$). In particular, when $\L(M_*) = O(b/N)$, the convergence rate given in Theorem~\ref{thm:mini-batch} is on the order of $O(b/N)$ while the convergence rate in (\ref{eqn:minibound}) is only $O(1/\sqrt{N})$.

\subsection{Adaptive Sampling based SGD for DML (AS-SGD)}
%%%%%%%%%%%%%%%%%%%%%%%%%%%%%%%%%%%%%%%%%%%%%%%%%%%%%%%%%%%%%%%%%%
%adaptive sampling
%%%%%%%%%%%%%%%%%%%%%%%%%%%%%%%%%%%%%%%%%%%%%%%%%%%%%%%%%%%%%%%%%%
\begin{algorithm}[t]
\caption{Adaptive Sampling Stochastic Gradient Descent (AS-SGD) for DML}
\begin{algorithmic}[1]
\STATE {\bf Input:} triplet constraints $\{(\x_i^t, \x_j^t, \x_k^t)\}_{t=1}^N$, step size $\eta$, and domain size $R$

\STATE Initialize $M_1 = I$

\FOR{$t=1, \ldots, N$}
    \STATE Sample a binary random variable $Z_t$ with
    \[
    \Pr(Z_t = 1) = |\ell'(\Delta(\x_i^t, \x_j^t, \x_k^t; M_t)|
    \]
    \IF{$Z_t = 1$}
        \STATE Update the distance metric by
        \begin{eqnarray*}
            \tau_t & = & \rm{sign}(\ell'(\Delta(\x_i^t, \x_j^t, \x_k^t; M_t))\\
            M_{t+1} & = & \Pi_{\Omega}\left(M_t - \eta \tau_tA_t\right)
           % M_{t+1} & = & \frac{1}{\max(\|\Mh_{t+1}\|_F/R, 1)} \Mh_{t+1}
        \end{eqnarray*}
    \ENDIF
\ENDFOR
\RETURN $\bar{M} = \frac{1}{N}\sum_{t=1}^{N} M_{t}$
\end{algorithmic}\label{alg:2}
\end{algorithm}

We now develop a new approach for reducing the number of updates in SGD in order to improve the computational efficiency of DML. Instead of updating the distance metric at each iteration, the proposed strategy introduces a random binary variable to decide if the distance metric $M_t$ will be updated given a triplet constraint $(\x_i^t, \x_j^t, \x_k^t)$. More specifically, it computes the derivative $\ell'(\Delta(\x_i^t, \x_j^t, \x_k^t; M_t))$, and samples a random variable $Z_t$ with probability
\[
\Pr(Z_t = 1) = |\ell'(\Delta(\x_i^t, \x_j^t, \x_k^t; M_t))|
\]
The distance metric will be updated only when $Z_t = 1$. According to Proposition 2, we have $|\ell'(\Delta(\x_i^t, \x_j^t, \x_k^t; M_t))| \leq L \ell(\Delta(\x_i^t, \x_j^t, \x_k^t; M_t))$ for the smooth loss function given in~(\ref{eqn:smooth}), implying that a triplet constraint has a high chance to be used for updating the distance metric if it has a large loss. Therefore, the essential idea of the proposed adaptive sampling strategy is to give a large chance to update the distance metric when the triplet is difficult to be classified and a low chance when the triplet can be classified correctly with large margin. We note that an alternative strategy is to sample a triplet constraint $(\x_i^t, \x_j^t, \x_k^t)$ base on its loss $\ell(\Delta(\x_i^t, \x_j^t, \x_k^t; M_t))$. We did not choose the loss as the basis for updating because it is the derivative, not the loss, that will be used by SGD for updating the distance metric. The detailed steps of adaptive sampling based SGD for DML is given in Algorithm~\ref{alg:2}. We refer to this algorithm as {\bf AS-SGD} for short in the rest of this paper.

The theorem below provides the performance guarantee for AS-SGD. It also bounds the number of updates $\sum_{t=1}^T Z_t$ for AS-SGD.
\begin{thm} \label{thm:as}
Let $\bar{M}$ be the solution output by Algorithm~\ref{alg:2} that uses the loss function defined in (\ref{eqn:smooth}). Let $M_*$ be the optimal solution to (\ref{eqn:opt}). Assume $\|A_t\|_F \leq A$ for any triplet constraint. For a fixed $\delta \in (0, 1)$, %assuming $N \geq 9\ln(1/\delta)$,
we have, with a probability $1 - 2\delta$:
\begin{eqnarray}
\L(\bar{M})\leq\frac{\L(M_*)}{1-3\eta LA^2} + \frac{C_2}{(1 - 3\eta LA^2)N}\left(\frac{R^2}{\eta}+ \eta + 1\right) \label{eqn:as}
\end{eqnarray}
and
\begin{eqnarray}
    \sum_{t=1}^N Z_t \leq \frac{3}{2}L\sum_{t=1}^N\ell(M_t) + \frac{5}{2}\ln{\frac{m}{\delta}} \label{eqn:update-bound}
\end{eqnarray}
where
\begin{eqnarray*}
C_2 & = & \max\left\{\frac{1}{2}+16\ln{\frac{m}{\delta}},\frac{5}{4}A^2\ln{\frac{m}{\delta}},RA\ln{\frac{2 m}{\delta}}\right\}\\
m   & = & \lceil \log_2 (N^2) \rceil
\end{eqnarray*}
\end{thm}

\begin{figure*}[!ht]
\centering
\begin{minipage}[h]{2.2in}
\centering
\includegraphics[width= 2.3in ]{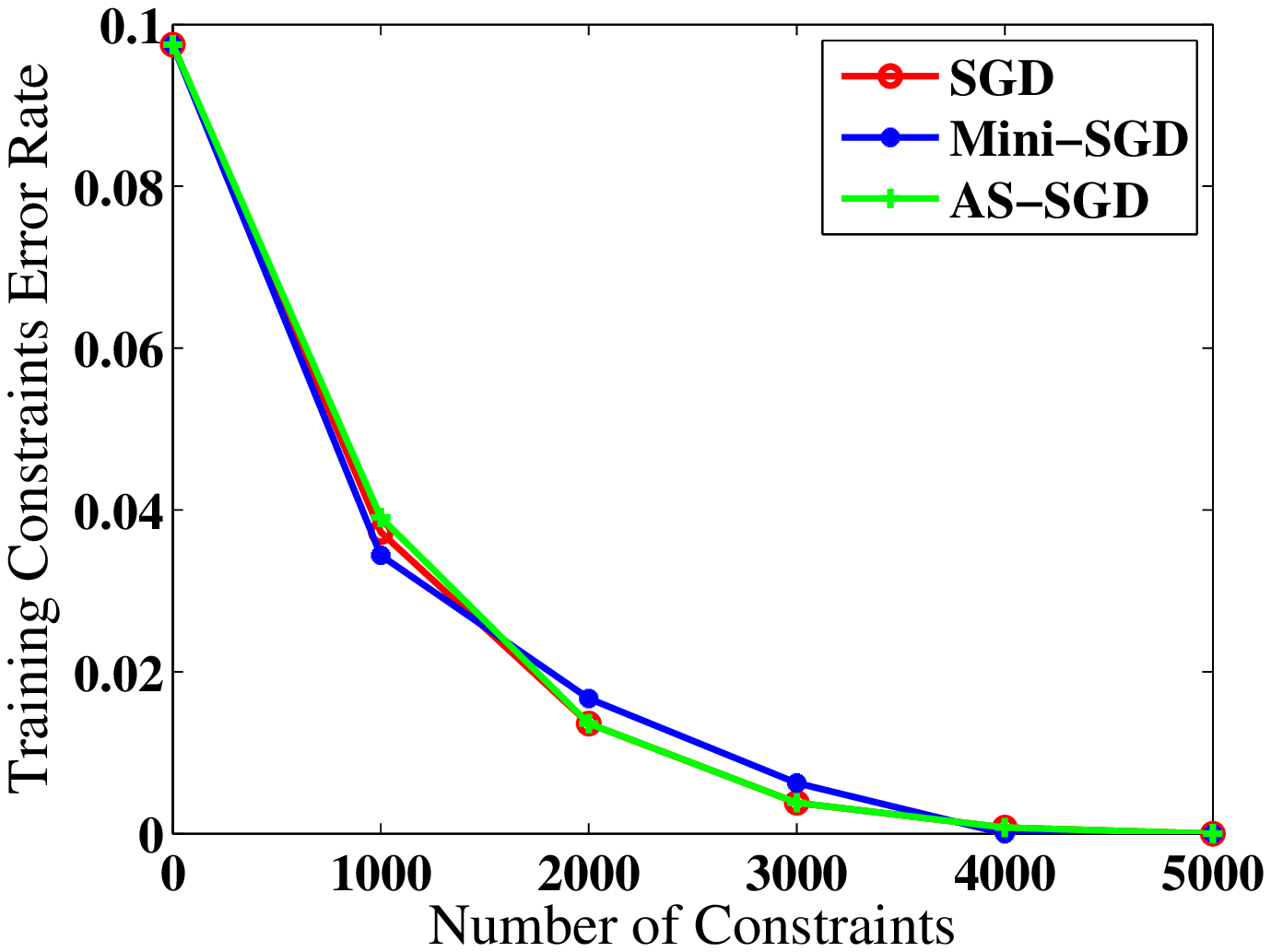}\\
\mbox{\footnotesize (a) {\it semeion}}
\end{minipage}
\begin{minipage}[h]{2.2in}
\centering
\includegraphics[width= 2.3in ]{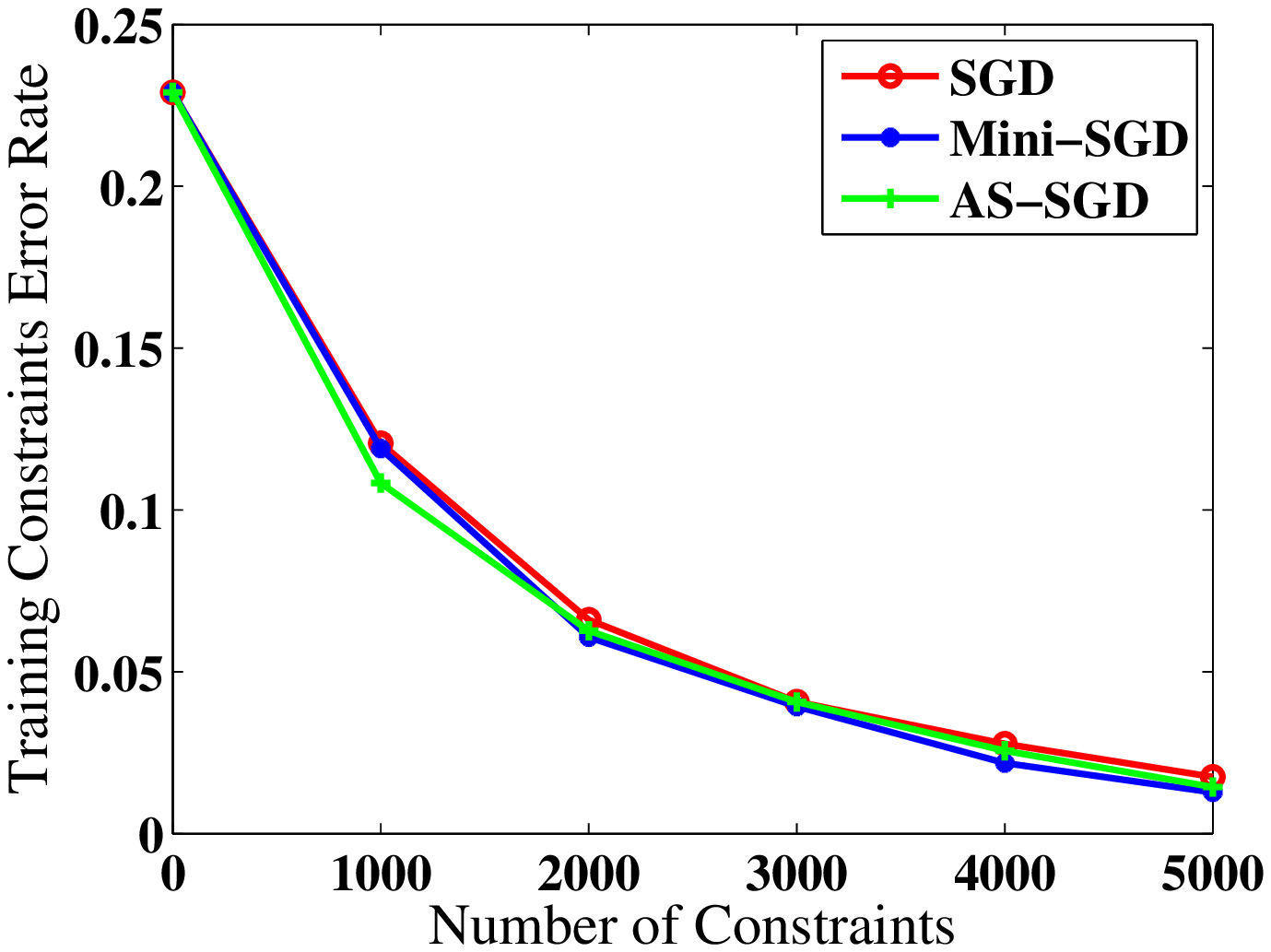}\\
\mbox{\footnotesize (b) {\it dna}}
\end{minipage}
\begin{minipage}[h]{2.2in}
\centering
\includegraphics[width= 2.3in ]{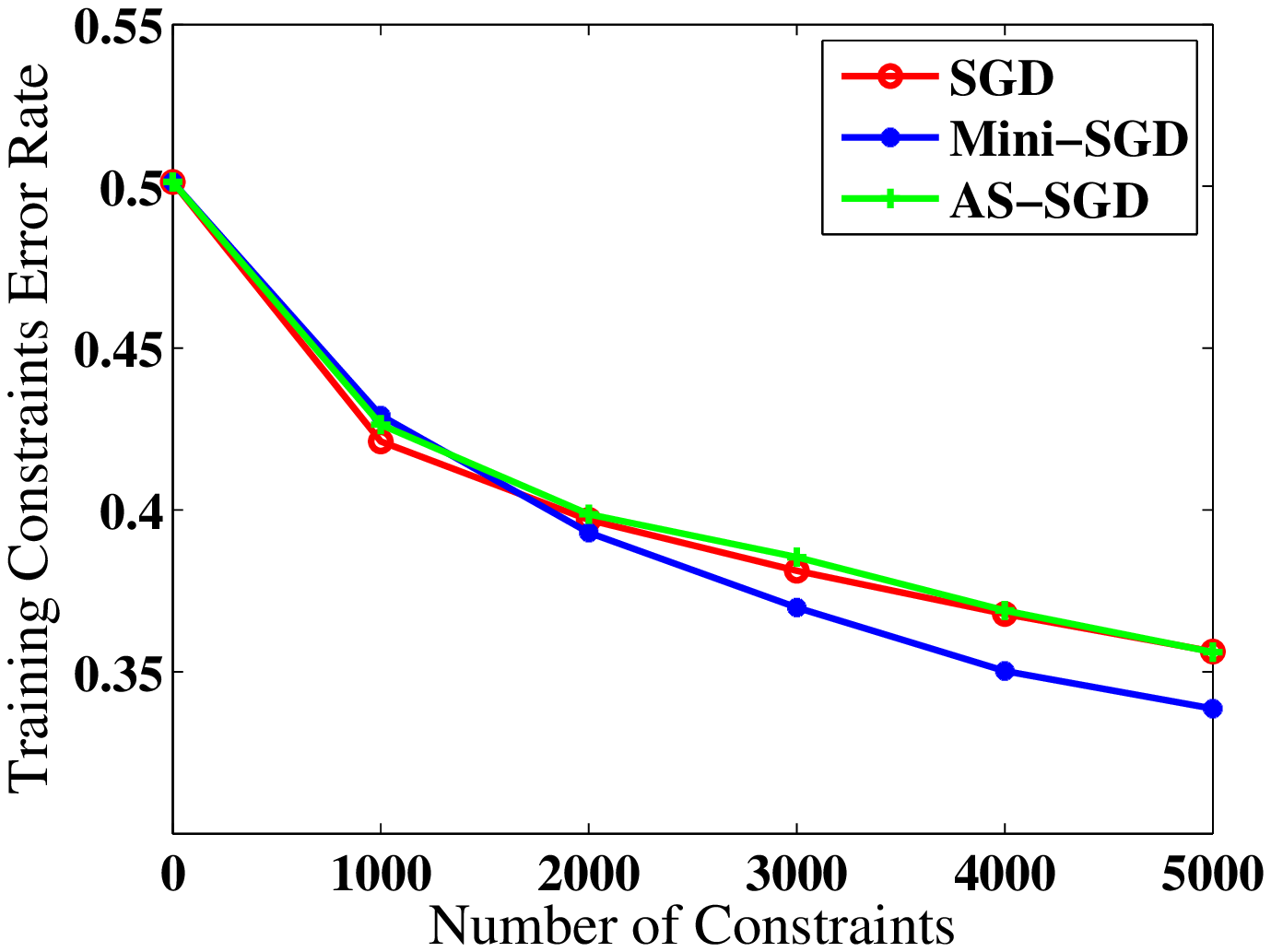}\\
\mbox{\footnotesize (c) {\it protein}}
\end{minipage}
\caption{The convergence of different SGD algorithms}\label{fig1}
\end{figure*}

\noindent{\bf Remark 2} The bound given in (\ref{eqn:as}) shares similar structure as that given in (\ref{eqn:mini-bound}) except that it does not have mini-batch size $b$ that can be used to make tradeoff between the number of updates and the classification accuracy. The number of updates performed by Algorithm~\ref{alg:2} is bounded by (\ref{eqn:update-bound}). The dominate term in (\ref{eqn:update-bound}) is $O(\sum_{t=1}^N\ell(M_t))$, implying that Algorithm~\ref{alg:2} will have a small number of updates if the learned distance metric $M_t$ can classify the triplet constraint correctly at most iterations. In other words, the smaller the number of classification mistakes made by the learned distance metric $M_t$, the less number of updates will be performed by Algorithm~\ref{alg:2}. We validate the theorem by running the AS-SGD algorithm on three datasets. Figure~\ref{fig1} shows the reduction in the training error over the number of triplet constraints by AS-SGD and the standard SGD algorithm. We observe that AS-SGD converges to a similar value of training error as the full SGD algorithm.

\subsection{Hybrid Approaches: Combine Mini-batch with Adaptive Sampling for DML}

%%%%%%%%%%%%%%%%%%%%%%%%%%%%%%%%%%%%%%%%%%%%%%%%%%%%%%%%%%%%%%%%%%
%mini-batch with adaptive sampling
%%%%%%%%%%%%%%%%%%%%%%%%%%%%%%%%%%%%%%%%%%%%%%%%%%%%%%%%%%%%%%%%%%
\begin{algorithm}[t]
\caption{A Framework of Hybrid Stochastic Gradient Descent (Hybrid-SGD) for DML}
\begin{algorithmic}[1]

\STATE {\bf Input:} triplet constraints $\{(\x_i^t, \x_j^t, \x_k^t)\}_{t=1}^N$, step size $\eta$, mini-batch size $b$, and domain size $R$

\STATE Initialize $M_1 = I$ and $T = N/b$

\FOR{$t=1, \ldots, T$}
    \STATE Sample $b$ triplets $\{\x_i^{t,s}, \x_j^{t,s}, \x_k^{t,s}\}_{s=1}^b$.
    \STATE \framebox{Compute sampling probability $\gamma_t$.}
    \STATE Sample a binary random variable $Z_t$ with
    \[
        \Pr(Z_t = 1) = \gamma_t
    \]
    \IF{$Z_t = 1$}
    \STATE Update the distance metric by
        \begin{eqnarray*}
            \tau_t & = & 1/\gamma_t\\
            M_{t+1} & = & \Pi_{\Omega}(M_t - \eta \tau_t \nabla \ell_t(M_t))
           % M_{t+1}&=& \frac{1}{\max(\|\Mh_{t+1}\|_F/R,1)} \Mh_{t+1}
        \end{eqnarray*}
     \ENDIF
\ENDFOR
\RETURN $\bar{M} = \frac{1}{T}\sum_{t=1}^{T} M_{t}$
\end{algorithmic}\label{alg:3}
\end{algorithm}

Since mini-batch and adaptive sampling improve the computational efficiency of SGD from different aspects, it is natural to combine them together for more efficient DML. Similar to the Mini-SGD algorithm, the hybrid approaches will group multiple triplet constraints into a mini-batch. But, unlike Mini-SGD that updates the distance metric for every mini-batch of constraints, the hybrid approaches follow the idea of adaptive sampling, and introduce a binary random variable to decide if the distance metric will be updated for every mini-batch of constraints. By combining the strength of mini-batch and adaptive sampling for SGD, the hybrid approaches are able to make further improvement in the computational efficiency of DML. Algorithm~\ref{alg:3} highlights the key steps of the hybrid approaches.

One of the key steps in the hybrid approaches (step 5 in Algorithm~\ref{alg:3}) is to choose appropriate sampling probability $\gamma_t$ for every mini-batch constraints $(\x_i^{t,s}, \x_j^{t,s}, \x_k^{t,s}), s=1, \ldots, b$. In this work, we study two different choices for sampling probability $\gamma_t$:
\begin{itemize}
\item The first approach chooses $\gamma_t$ based on a triplet constraint randomly sampled from a mini-batch. More specifically, given a mini-batch of triplet constraints $\{\x_i^{t,s}, \x_j^{t,s}, \x_k^{t,s}\}_{s=1}^b$, it randomly samples an index $s'$ in the range $[1, b]$. It then sets the sampling probability $\gamma_t$ to be the derivative for the randomly sampled triplet, i.e.,
    \begin{equation*}
        \gamma_t = |\ell'(\Delta(\x_i^{t,s'}, \x_j^{t,s'}, \x_k^{t,s'}; M_t))|
    \end{equation*}
    We refer to this approach as {\bf HR-SGD}.
\item The second approach is based on the average case analysis. It sets the sampling probability as the average derivative measured by the norm of the gradient $\nabla \ell_t(M_t)$, i.e.,
    \begin{equation*}
        \gamma_t = \frac{1}{W}\|\nabla\ell_t(M_t)\|_F
    \end{equation*}
    where $W = \max_t \|\nabla\ell_t(M_t)\|_F$ and is estimated by sampling. We refer to this approach as {\bf HA-SGD}.
\end{itemize}

\section{Experiments}
\label{sec:exp}

\begin{table}[t]
\centering
\caption{Statistics for the ten datasets used in our empirical study.}
\label{rlr}
%\vspace{3mm}
\begin{tabular}{|c|c|c|c|c|}
\hline
         &\# class &\# feature&\# train &\# test\\\hline
semeion  &10  &256 &1,115   &478         \\\hline
dna      &3   &180 &2,000   &1,186      \\\hline
isolet   &26  &617 &6,238   &1,559      \\\hline
tdt30    &30  &200 &6,575   &2,819       \\\hline
letter   &26  &16  &15,000  &5,000      \\\hline
protein  &3   &357 &17,766  &6,621      \\\hline
connect4 &3   &42  &47,289  &20,268     \\\hline
sensit   &3   &100 &78,823  &19,705     \\\hline
rcv20    &20  &200 &477,141 &14,185   \\\hline
poker    &10  &10  &1,000,000 &25,010   \\\hline
\end{tabular}
\end{table}
%%%%%%%%%%%%%%%%%%%%%%%%%%%%%%%%%%%%%%%%%%%%%%%%%%%%%%%%
%dataset description
%%%%%%%%%%%%%%%%%%%%%%%%%%%%%%%%%%%%%%%%%%%%%%%%%%%%%%%%
Ten datasets are used to validate the effectiveness of the proposed algorithms. Table~\ref{rlr} summarizes the information of these datasets. Datasets {\it dna}, {\it letter}~\cite{HsuL02}, {\it protein} and {\it sensit}~\cite{DuarteH04} are downloaded from LIBSVM~\cite{ChangL11}. Datasets {\it tdt30} and {\it rcv20} are document corpora: {\it tdt30} is the subset of tdt2 data~\cite{CaiWH09} comprised of the documents from the $30$ most popular categories and {\it rcv20} is the subset of a large rcv1 dataset~\cite{BekkermanS08} consisted of documents from the $20$ most popular categories. We reduce the dimensionality of these document datasets to $200$ by principle components analysis (PCA). All the other datasets are downloaded directly from the UCI repository~\cite{Frank2010}. For most datasets used in this study, we use the standard training/testing split provided by the original dataset, except for datasets {\it semeion}, {\it connect4} and {\it tdt30}. For these three datasets, we randomly select $70\%$ of data for training and use the remaining $30\%$ for testing; experiments related to these three datasets are repeated ten times, and the prediction result averaged over ten trials is reported. All experiments are implemented on a laptop with 8GB memory and two 2.50GHz Intel Core i5-2520M CPUs.

\subsection{Parameter Setting}
%%%%%%%%%%%%%%%%%%%%%%%%%%%%%%%%%%%%%%%%%%%%%%%%%%%%%%%%
%influence of parameters
%%%%%%%%%%%%%%%%%%%%%%%%%%%%%%%%%%%%%%%%%%%%%%%%%%%%%%%%
The parameter $L$ in the loss function (\ref{eqn:smooth}) is set to be $3$ according to the suggestion in~\cite{zhang2001}. We set $N =100,000$ for the number of iterations (i.e., the number of triplet constraints). To construct a triplet constraint at each iteration $t$, we first randomly sample an example $(\x_i^t, y_i^t)$ from the training data; we then find two of its nearest neighbors $\x_j^t$ and $\x_k^t$, measured by Euclidean distance, from the training examples, with $\x_j^t$ sharing the same class label as $\x_i^t$ and $\x_k^t$ belonging to a class different from $y_i^t$. For Mini-SGD and the hybrid approaches, we set $b = 10$ for the size of mini-batch as in~\cite{shaw2011}, leading to a total of $T = 10,000$ iterations for these approaches. We evaluate the learned distance metric by the classification error of a $k$-NN on the test data, where the number of nearest neighbors $k$ is set to be $3$ based on our experience.

Parameter $R$ in the proposed algorithms determines the domain size for the distance metric to be learned. We observe that the classification error of $k$-NN remains almost unchanged when varying $R$ in the range of $\{100, 1000, 10000\}$. We thus set $R = 1,000$ for all the experiments. Another important parameter used by the proposed algorithms is the step size $\eta$. We evaluate the impact of step size $\eta$ by measuring the classification error of a $k$-NN algorithm that uses the distance metric learned by the Mini-SGD algorithm with $\eta = \{0.1,1,10\}$. We observe that $\eta = 1$ yields a low classification error for almost all datasets by cross-validation with $R = 1,000$ and $T = 10$. We thus fix $\eta = 1$ for the proposed algorithms in all the experiments.

\subsection{Experiment (I): Effectiveness of the Proposed SGD Algorithms for DML}
In this experiment, we compare the performance of the proposed SGD algorithms for DML, i.e., Mini-SGD, AS-SGD and two hybrid approaches (HR-SGD and HA-SGD), to the full version of SGD for DML (SGD). We also include Euclidean distance as the reference method in our comparison. Table~\ref{rlr4} shows the classification error of $k$-NN ($k=3$) using the distance metric learned by different DML algorithms. First, it is not surprising to observe that all the distance metric learning algorithms improve the classification performance of $k$-NN compared to the Euclidean distance. Second, for almost all datasets, we observe that all the proposed DML algorithms (i.e., Mini-SGD, AS-SGD, HR-SGD, and HA-SGD) yield similar classification performance as SGD, the full version of SGD algorithm for DML. This result confirms that the proposed SGD algorithms are effective for DML despite the modifications we made to the SGD algorithm.

\subsection{Experiment (II): Efficiency of the Proposed SGD Algorithms for DML}

Table.~\ref{rlr5} summarizes the running time for the proposed DML algorithms and the SGD method. We note that the running time in Table~\ref{rlr5} does not take into account the time for constructing triplet constraints since it is shared by all the methods in comparison.

It is not surprising to observe that all the proposed SGD algorithms, including Mini-SGD, AS-SGD, HA-SGD and HR-SGD, significantly reduce the running time of SGD. For instance, for dataset {\it isolet}, it takes SGD more than $32,000$ seconds to learn a distance metric, while the running time is reduced to less than $3,500$ seconds when applying the proposed SGD algorithms, roughly a factor of $10$ reduction in running time. Comparing the running time of AS-SGD to that of Mini-SGD, we observe that each method has its own advantage: AS-SGD is more efficient on datasets {\it semeion}, {\it dna}, {\it isolet}, and {\it tdt30}, while Mini-SGD is more efficient on the other six datasets. This is because different mechanisms are employed by AS-SGD and Mini-SGD to reduce the computational cost: AS-SGD improves the computational efficiency of DML by skipping the constraints that are easy to be classified, while Mini-SGD improves the the computational efficiency of SGD by performing the updating of distance metric once for multiple triplet constraints. Finally, we observe that the two hybrid approaches that combine the strength of both adaptive sampling and mini-batch SGD, are computationally most efficient for almost all datasets. We also observe that HR-SGD appears to be more efficient than HA-SGD on six datasets and only loses on datasets {\it protein}, {\it sensit} and {\it rcv20}. This is because HR-SGD computes the sampling probability $\gamma_t$ based on one randomly sampled triplet while HA-SGD needs to compute the average derivative for each mini-batch of triplet constraints for the sampling probability.

To further examine the computational efficiency of proposed SGD algorithms for DML, we summarize in Table~\ref{rlr6} the number of updating performed by different SGD algorithms. We observe that all the proposed SGD algorithms for DML are able to reduce the number of updates significantly compared to SGD. Comparing Mini-SGD to AS-SGD, we observe that for some datasets (e.g., {\it semeion}, {\it dna}, {\it isolet}, and {\it tdt30}), the number of updates performed by AS-SGD is significantly less than Mini-SGD, while it is the other way around for the other datasets. This is again due to the fact that AS-SGD and Mini-SGD deploy different mechanisms for reducing computational costs. As we expect, the two hybrid approaches are able to further reduce the number of updates performed by AS-SGD and Mini-SGD, making them more efficient algorithms for DML.

%%%%%%%%%%%%%%%%%%%%%%%%%%%%%%%%%%
%Update part
%%%%%%%%%%%%%%%%%%%%%%%%%%%%%%%%%%
By comparing the results in Table~\ref{rlr5} to the results in Table~\ref{rlr6}, we observe that a small number of updates does NOT always guarantee a short running time. This is exhibited by the comparison between the two hybrid approaches: although HA-SGD performs the similar number of updates as HR-SGD on datasets {\it dna} and {\it isolet}, it takes HA-SGD significantly longer time to finish the computation than HR-SGD. This is also exhibited by comparing the results across different datasets for a fixed method. For example, for the HA-SGD method, the number of updates for the {\it protein} dataset is nearly the same as that for the {\it poker} dataset, but the running time for the {\it protein} dataset is about $50$ times longer than that for the {\it poker} dataset. This result may sound counter intuitive at the first glance. But, a more careful analysis reveals that in addition to the number of updates, the running time of DML is also affected by the computational cost per iteration, which explains the consistency between Table~\ref{rlr5} and \ref{rlr6}. In the case of comparing the two hybrid approaches, we observe that HA-SGD is subjected to a higher computational cost per iteration than HR-SGD because HA-SGD has to compute the norm of the {\it average} gradient over each mini-batch while HR-SGD only needs to compute the derivative of {\it one} randomly sampled triplet constraint for each mini-batch. In the case of comparing the running time across different datasets, the {\it protein} dataset has a significantly higher dimensionality than the {\it poker} dataset, and therefore is subjected to a higher computational cost per iteration because the computational cost of projecting an updated distance metric onto the PSD cone increases at least quadratically in the dimensionality.

\begin{table*}[t]
\centering
\caption{Classification error ($\%$) of $k$-NN ($k=3$) using the distance metrics learned by different SGD methods, online learning algorithms and batch learning approach for DML.}\label{rlr4}
\begin{tabular}{|c|c|c|c|c|c|c||c|c|c|c|}
\hline
          &Baseline&Batch&\multicolumn{3}{c|}{Online Learning}& &\multicolumn{4}{c|}{Proposed Methods} \\\hline
          &Euclidean&LMNN  &LEGO   &OASIS &SPML  &\ \ SGD\ \  &Mini-SGD  &AS-SGD &HR-SGD &HA-SGD   \\\hline
semeion   &8.7      &9.0   &11.9   &8.3   &6.3   &6.3  &6.5       &6.3   &6.4       &6.2 \\\hline
dna       &20.7     &6.2   &9.3    &16.6  &9.1   &8.6  &9.4       &8.4   &8.1       &8.1\\\hline
isolet    &9.0      &5.4   &8.3    &6.5   &6.6   &6.3  &6.2       &6.0   &6.4       &6.1  \\\hline
tdt30     &5.3      &3.0   &14.6   &4.0   &3.7   &3.8  &3.7       &3.7   &3.8       &3.6 \\\hline
letter    &4.4      &3.2   &4.0    &2.2   &3.1   &2.1  &2.5       &2.1   &2.5       &2.3  \\\hline
protein   &50.0     &40.1  &42.4   &40.1  &41.9  &40.7 &38.9      &40.7  &41.0      &40.9  \\\hline
connect4  &29.5     &21.1  &25.8   &22.1  &24.5  &20.1 &20.1      &20.1  &22.2      &20.4  \\\hline
sensit    &27.3     &24.3  &25.4   &24.1  &23.7  &24.0 &24.0      &24.0  &24.4      &24.6  \\\hline
rcv20     &9.1      &N/A   &8.9    &8.6   &8.9   &8.5  &8.7       &8.4   &8.4       &8.6  \\\hline
poker     &38.0     &N/A   &39.2   &36.1  &37.8  &35.0 &33.8      &35.0  &34.3      &34.4  \\\hline
\end{tabular}
\end{table*}

%%%%%%%%%%%%%%%%%%%%%%%%%%%%%%%%%%%%%%%%%%%%%%%%%%%%%%%%
%running time
%%%%%%%%%%%%%%%%%%%%%%%%%%%%%%%%%%%%%%%%%%%%%%%%%%%%%%%%

\begin{table*}[t]
\centering
\caption{Running time (seconds) for different SGD methods, online learning algorithms and batch learning approach for DML. Note that LMNN, a batch DML algorithm, is mainly implemented in C, while the other algorithms in comparison are implemented in Matlab, which is usually less efficient than C.}\label{rlr5}
\begin{tabular}{|c|c|c|c|c|c||c|c|c|c|}
\hline
           &Batch &\multicolumn{3}{c|}{Online Learning}&&\multicolumn{4}{c|}{Proposed Methods} \\\hline
           &LMNN    &LEGO      &OASIS   &SPML     &SGD       &Mini-SGD &AS-SGD &HR-SGD  &HA-SGD  \\\hline
semeion    &112.7   &355.8     &29.1    &206.6    &2,172.4   &263.2    &45.2   &7.4      &42.4 \\\hline
dna        &255.9   &330.2     &39.1    &122.1    &1,165.3   &121.0    &30.6   &7.1      &28.0 \\\hline
isolet     &2,454.3 &3,454.2   &515.7   &3,017.2  &32,762.7  &3,440.7  &908.4  &127.6    &246.3 \\\hline
tdt30      &264.5   &372.6     &51.2    &145.1    &1,351.0   &148.0    &108.8  &11.6     &41.6 \\\hline
letter     &251.6   &15.0      &10.8    &5.6      &27.3      &5.3      &10.9   &1.8      &3.2 \\\hline
protein    &3,906.4 &1,318.9   &3,825.9 &573.8    &5,448.9   &580.6    &1,335.8&184.5    &145.6 \\\hline
connect4   &540.2   &23.1      &79.0    &16.4     &109.6     &15.9     &60.5   &8.0      &6.97  \\\hline
sensit     &10,481.2&93.3      &303.9   &44.3     &365.4     &41.3     &243.9  &26.2     &17.9  \\\hline
rcv20      &N/A     &443.6     &1,313.7 &154.4    &1,542.1   &158.4    &932.9  &101.4    &45.8  \\\hline
poker      &N/A     &17.3      &17.6    &5.8      &21.0      &4.5      &13.5   &2.8      &3.4  \\\hline
\end{tabular}
\end{table*}

\begin{table*}[t]
\centering
\caption{The number of updates for different SGD methods and online learning algorithms for DML.}\label{rlr6}\vspace{3mm}
\begin{tabular}{|c|c|c|c|c||c|c|c|c|}
\hline
          &\multicolumn{3}{|c|}{Online Learning}&&\multicolumn{4}{c|}{Proposed Methods} \\\hline
          &LEGO      &OASIS   &\ SPML\   &SGD    &Mini-SGD&AS-SGD  &HR-SGD  &HA-SGD  \\\hline
semeion   &71,142.4  &432.7  &10,000 &100,000&10,000&142.2    &101.4   &162.8  \\\hline
dna       &140,027   &2,042  &10,000 &100,000&10,000&707      &351     &372 \\\hline
isolet    &110,175   &1,426  &10,000 &100,000&10,000&1,893    &353     &378   \\\hline
tdt30     &131,997.6 &2,284.6&10,000 &100,000&10,000&5,563.7  &567.6   &784.6\\\hline
letter    &130,794   &28,063 &10,000 &100,000&10,000&12,931   &1,398   &457  \\\hline
protein   &166,384   &64,804 &10,000 &100,000&10,000&22,127   &3,064   &1,623   \\\hline
connect4  &153,311.6 &69,865 &10,000 &100,000&10,000&44,510.8 &4,161.2 &2,134.3 \\\hline
sensit    &162,869   &78,223 &10,000 &100,000&10,000&60,028   &5,675   &1,281   \\\hline
rcv20     &137,246   &88,476 &10,000 &100,000&10,000&60,708   &6,095   &779  \\\hline
poker     &179,714   &71,620 &10,000 &100,000&10,000&43,259   &4,111   &1,635  \\\hline
\end{tabular}
\end{table*}

\subsection{Experiment (III): Comparison with State-of-the-art Online DML Methods}
%%%%%%%%%%%%%%%%%%%%%%%%%%%%%%%%%%%%%%%%%%%%%%%%%%%%%%%%
%baseline description
%%%%%%%%%%%%%%%%%%%%%%%%%%%%%%%%%%%%%%%%%%%%%%%%%%%%%%%%
We compare the proposed SGD algorithms to three state-of-the-art online algorithms and one bath method for DML:
\begin{compactitem}
\item {\bf SPML}~\cite{shaw2011}: an online learning algorithm for DML that is based on mini-batch SGD and the hinge loss,
\item {\bf OASIS}~\cite{chechik2010}: a state-of-the-art online DML algorithm,
\item {\bf LEGO}~\cite{JainKDG08}: an online version of the information theoretic based DML algorithm~\cite{DavisKJSD07}.
\end{compactitem}

Finally, for sanity checking, we also compare the proposed SGD algorithms to {\bf LMNN}~\cite{weinberger2009}, a state-of-the-art batch learning algorithm for DML.

Both SPML and OASIS use the same set of triplet constraints to learn a distance metric as the proposed SGD algorithms. However, unlike SPML and OASIS, pairwise constraints are used by LEGO for DML. For fair comparison, we generate the pairwise constraints for LEGO by splitting each triplet constraint $(\x_i^t, \x_j^t, \x_k^t)$ into two pairwise constraints: a must-link constraint $(\x_i^t, \x_j^t)$ and a cannot-link constraint $(\x_i^t, \x_k^t)$. This splitting operation results in a total of $200,000$ pairwise constraints for LEGO. Finally, we note that since LMNN is a batch learning method, it is allowed to utilize {\it any} triplet constraint derived from the data, and is not restricted to the set of triplet constraints we generate for the SGD methods. All the baseline DML algorithms are implemented by using the codes from the original authors except for SPML, for which we made appropriate changes to the original code in order to avoid large matrix multiplication and improve the computational efficiency. SPML, OASIS and LEGO are implemented in Matlab, while the core parts of LMNN are implemented by C that is usually deemed to be more efficient than Matlab. The default parameters suggested by the original authors are used in the baseline algorithms. The step size of LEGO is set to be $1$, as it was observed in ~\cite{chechik2010} that the prediction performance of LEGO is in general insensitive to the step size. In all experiments, all the baseline methods set the initial solution for distance metric to be an identity matrix.

Table.~\ref{rlr4} summarizes the classification results of $k$-NN ($k=3$) using the distance metrics learned by the four baseline algorithms. First, we observe that LEGO performs significantly worse than the proposed DML algorithms for five datasets, including {\it semeion}, {\it isolet}, {\it tdt30}, {\it connect4}, and {\it poker}. This can be explained by the fact that LEGO uses pairwise constraints for DML while the other methods in comparison use triplet constraints for DML. According to~\cite{chechik2010,shaw2011,weinberger2009}, triplet constraints are in general more effective than pairwise constraints. Second, although both SPML and Mini-SGD are based on the mini-batch strategy, SPML performs significantly worse than Mini-SGD on three datasets, i.e. {\it protein}, {\it connect4}, and {\it poker}. The performance difference between SPML and Mini-SGD can be explained by the fact that Mini-SGD uses a smooth loss function while a hinge loss is used by SPML. According to our analysis and the analysis in~\cite{cotter2011}, using a smooth loss function is critical for the success of the mini-batch strategy. Third, OASIS yields similar performance as the proposed algorithms for almost all datasets except for datasets {\it semeion}, {\it dna} and {\it poker}, for which OASIS performs significantly worse. Overall, we conclude that the proposed DML algorithms yield similar, if not better, performance as the state-of-the-art online learning algorithms for DML.

Compared to LMNN, a state-of-the-art batch learning algorithm for DML, we observe that the proposed SGD algorithms yield similar performance on three datasets. They however perform significantly better than LMNN on datasets {\it semeion} and {\it letter}, and significantly worse on datasets {\it dna}, {\it isolet} and {\it tdt30}. We attribute the difference in classification error to the fact that the proposed DML algorithms are restricted to $100,000$ randomly sampled triplet constraints while LMNN is allowed to use {\it all} the triplet constraints that can be derived from the data. The restriction in triplet constraints could sometimes limit the classification performance but at the other time help avoid the overfitting problem. We also observe that LMNN is unable to run on the two large datasets {\it rcv20} and {\it poker}, indicating that LMNN does not scale well to the size of datasets.

The running time and the number of updates of the baseline online DML algorithms can be found in Table~\ref{rlr5} and Table~\ref{rlr6}, respectively. It is not surprising to observe that the three online DML algorithms are significantly more efficient than SGD in terms of both running time and the number of updates. We also observe that Mini-SGD and SPML share the same number of updates and similar running time for all datasets because they use the same mini-batch strategy. Furthermore, compared to the three online DML algorithms, the two hybrid approaches are significantly more efficient in both running time and the number of updates. Finally, since LMNN is implemented by C, it is not surprising to observe that LMNN shares similar running time as the other online DML algorithms for relatively small datasets. It is however significantly less efficient than the online learning algorithms for datasets of modest size (e.g. {\it connect4} and {\it sensit}), and becomes computationally infeasible for the two large datasets {\it rcv20} and {\it poker}. Overall, we observe that the two hybrid approaches are significantly more efficient than the other DML algorithms in comparison.

\section{Conclusion}
\label{sec:conclusion}

In this paper, we propose two strategies to improve the computational efficiency of SGD for DML, i.e. mini-batch and  adaptive sampling. The key idea of mini-batch is to group multiple triplet constraints into a mini-batch, and only update the distance metric once for each mini-batch; the key idea of adaptive sampling is to perform stochastic updating by giving a difficult triplet constraint more chance to be used for updating the distance metric than an easy triplet constraint. We develop theoretical guarantees for both strategies. We also develop two variants of hybrid approaches that combine mini-batch with adaptive sampling for more efficient DML. Our empirical study confirms that the proposed algorithms yield similar, if not better, prediction performance as the state-of-the-art online learning algorithms for DML but with significantly less amount of running time. Since our empirical study is currently limited to datasets with relatively small number of features, we plan to examine the effectiveness of the proposed algorithms for DML with high dimensional data.

\bibliographystyle{abbrv}
\bibliography{dist}  % sigproc.bib is the name of the Bibliography in this case

\appendix

The analysis for Theorem~\ref{thm:mini-batch} is in the supplementary document~\footnote{https://sites.google.com/site/zljzju/Supplymentary.pdf} and we give the proof for Theorem~\ref{thm:as} here. Define:
\begin{equation*}
\begin{array}{l}
\begin{array}{ll}
C_N = \sum_{t=1}^N |\ell'(M_t)|, & X_t = Z_t - |\ell'(M_t)|,\\
\Lambda _N = \sum_{1\leq t\leq N}^N X_t, & K = \max\limits_{1 \leq t \leq N} X_t \leq 1,
\end{array}\\
\sigma_N^2 = \sum_{t=1}^N E[(Z_t - |\ell'(M_t)|)^2]\leq\sum_{t=1}^N |\ell'(M_t)| = C_N
\end{array}
\end{equation*}
Using Berstein inequality for martingales~\cite{nicolo2006}, we have:
\begin{eqnarray*}
&&\Pr(\Lambda_N\geq 2\sqrt{C_N\tau}+\sqrt{2}K\tau/3)\\
&=&\Pr(\Lambda_N\geq 2\sqrt{C_N\tau}+\sqrt{2}K\tau/3,\sigma_N^2\leq C_N,C_N\leq N)\\
&\leq&\Pr\left(\begin{array}{l}\Lambda_N\geq 2\sqrt{C_N\tau}+\sqrt{2}K\tau/3,\sigma_N^2\leq C_N,\\C_N\leq 1/N\end{array}\right)\\
&&+\sum_{i=1}^m\Pr\left(\begin{array}{l}\Lambda_N\geq 2\sqrt{C_N\tau}+\sqrt{2}K\tau/3,\sigma_N^2\leq C_N,\\2^{i-1}/N<C_N\leq2^i/N\end{array}\right)\\
&\leq&\Pr(C_N\leq 1/N)\\
&&+ \sum_{i=1}^m\Pr\left(\Lambda_N\geq \sqrt{2\frac{2^i}{N}\tau}+\sqrt{2}K\tau/3,\sigma_N^2\leq \frac{2^i}{N}\right)\\
&\leq&\Pr(C_N\leq 1/N)+me^{-\tau}
\end{eqnarray*}
where $m = \lceil \log_2(N^2)\rceil$. By setting $me^{-\tau} = \delta$, with a probability $1-\delta$, the number of updates can be bounded as:
\begin{eqnarray}\label{eqn:up-bound}
\sum_{t=1}^N Z_t &\leq& C_N+\frac{1}{2}C_N+2\ln{\frac{m}{\delta}}+\frac{\sqrt{2}}{3}K\ln{\frac{m}{\delta}}\nonumber\\
&\leq& \frac{3}{2}L\sum_{t=1}^N\ell(M_t)+\frac{5}{2}\ln{\frac{m}{\delta}}
\end{eqnarray}

Then, we give the regret bound. Using the standard analysis for online learning~\cite{nicolo2006}, we have:
\begin{eqnarray*}
&&\ell(M_t)-\ell(M_*)\leq \langle\ell'(M_t)A_t,M_t-M_*\rangle\\
&=& \tau_tZ_t\langle A_t,M_t-M_*\rangle\\
&&+(\ell'(M_t)-\tau_tZ_t)\langle A_t,M_t-M_*\rangle\\
&\leq& \frac{\|M_t-M_*\|_F^2-\|M_{t+1}-M_*\|_F^2}{2\eta}+\frac{\eta A^2 Z_t}{2}\\
&&+\tau_t(|\ell'(M_t)|-Z_t)\langle A_t,M_t-M_*\rangle
\end{eqnarray*}

Taking the sum from $t=1$ to $N$, we have:
\begin{eqnarray*}
&&\sum_{t=1}^N \ell(M_t)-\ell(M_*)\leq\frac{\|M_1-M_*\|_F^2}{2\eta}+\frac{\eta A^2}{2}\sum_{t=1}^NZ_t\\
&&+\sum_{t=1}^N2\tau_t(|\ell'(M_t)|-Z_t)RA
\end{eqnarray*}
According to (\ref{eqn:up-bound}), with a probability $1-\delta$, the second item could be bounded as:
\begin{eqnarray}\label{eqn:second}
\frac{\eta A^2}{2}\sum_{t=1}^NZ_t&\leq&\eta A^2(\frac{3}{4}L\sum_{t=1}^N\ell(M_t)+\frac{5}{4}\ln{\frac{m}{\delta}})\nonumber\\
&\leq& \frac{3}{4}\gamma\sum_{t=1}^N \ell(M_t)+\frac{5}{4}\eta A^2\ln{\frac{m}{\delta}}
\end{eqnarray}
where $\gamma\geq \eta L A^2$.

Applying Berstein inequality for martingales~\cite{nicolo2006} for the last item, we have, with a probability $1-\delta$:
\begin{eqnarray*}
\begin{array}{l}
\sum_{t=1}^N 2\tau_t(|\ell'(M_t)|-Z_t)RA \leq 4RA\sqrt{C_N\ln{\frac{m}{\delta}}}+\frac{2\sqrt{2}}{3}RA\ln{\frac{m}{\delta}}
\end{array}
\end{eqnarray*}
\begin{eqnarray}\label{eqn:last}
&\leq& \frac{\gamma}{4}\sum_{t=1}^N\ell(M_t)+\frac{16R^2}{\eta}\ln{\frac{m}{\delta}}+RA\ln{\frac{m}{\delta}}
\end{eqnarray}

Combining the bounds in (\ref{eqn:second}) and (\ref{eqn:last}), we have, with a probability $1 - 2\delta$:
\begin{eqnarray*}
&&\sum_{t=1}^N \ell(M_t)- \ell(M_*)\leq \frac{1}{2\eta}(R^2+32R^2\ln{\frac{m}{\delta}})\\
&&+\gamma\sum_{t=1}^N\ell(M_t)+\frac{5}{4}\eta A^2\ln{\frac{m}{\delta}}+RA\ln{\frac{m}{\delta}}
\end{eqnarray*}
which is equal to:
\begin{eqnarray*}
\L(\bar{M})\leq\frac{1}{1-\gamma}(\L(M_*)+\frac{R^2c}{\eta N}+\frac{\eta c}{N}+\frac{c}{N})
\end{eqnarray*}
where
\[c = \max\left\{\frac{1}{2}+16\ln{\frac{m}{\delta}},\frac{5}{4}A^2\ln{\frac{m}{\delta}},RA\ln{\frac{m}{\delta}}\right\}\]
The proof is completed by setting $\gamma = 3\eta LA^2$.

\end{document}